\documentclass[sigconf]{acmart}
\usepackage{amsmath}
\usepackage{pbox}
\usepackage{pdflscape}
\usepackage{tabularx}
\usepackage{wrapfig}

\usepackage{amsmath}

\AtBeginDocument{%
  \providecommand\BibTeX{{%
    \normalfont B\kern-0.5em{\scshape i\kern-0.25em b}\kern-0.8em\TeX}}}

\renewcommand\footnotetextcopyrightpermission[1]{}

\acmConference[DLP-KDD 2022]{}{August 14, 2022}{Washington, DC, USA}

\begin{document}

\title{Self-supervised Learning for Hyperspectral Images of Trees}

%

\author{{{Moqsadur Rahman}$^1$}, {{Saurav Kumar}$^2$}, Santosh S. Palmate$^3$, M. Shahriar Hossain$^1$}
\affiliation{%
	\institution{$^1$\textit{Department of Computer Science, University of Texas at El Paso, El Paso, TX 79968, \country{USA}}}
	\institution{$^2$\textit{School of Sustainable Engineering and Built Environment, Arizona State University, Tempe, AZ 85281, \country{USA}}}
	\institution{$^3$\textit{Texas A\&M AgriLife Research, Texas A\&M University, El Paso, TX 79927, \country{USA}}}
    \institution{\textit{mrahman20@miners.utep.edu, sk2@asu.edu, santosh.palmate@ag.tamu.edu, mhossain@utep.edu}}
}

%
\renewcommand{\shortauthors}{M. Rahman, S. Kumar, S. S. Palmate, and M. S. Hossain}

%
\begin{abstract}
Aerial remote sensing using multispectral and RGB imagers has provided a critical impetus to precision agriculture. Analysis of the hyperspectral images with limited or no labels is challenging. This paper focuses on self-supervised learning to create neural network embeddings reflecting vegetation properties of trees from aerial hyperspectral images of crop fields. Experimental results demonstrate that a constructed tree representation, using a vegetation property-related embedding space, performs better in downstream machine learning tasks compared to the direct use of hyperspectral vegetation properties as tree representations.
\end{abstract}




\maketitle
\section{Introduction}
There is a growing recognition that hyperspectral imaging can have transformative effects on plant stress, and disease diagnosis \cite{disease_1, disease_2, stress_1, stress_2}. Many plant stresses are difficult to identify using RGB images but are apparent in the non-visible spectrum making hyperspectral imaging a great choice for the detection of stress. Most of the past studies with hyperspectral images have looked at designing stress indicators and correlating the spectrum of the images to detect specific types of stress or disease. While the methods have worked well for correlating stresses and diseases with the spectrum, \textit{early detection} of these stresses and diseases is still hard to achieve \cite{stress_1}. Known patterns help detect stresses early. For example, drought stress \cite{stress_1, stress_2} and diseases that start in a small region on the foliage \cite{disease_2} can now be detected early using hyperspectral images.

Many of the stress patterns for early detection are still unknown. As a result, labeled data to classify stresses are not widely available. Moreover, there are myriads of relationships between hyperspectral bandwidths that could help detect stresses and diseases early. This paper focuses on designing a self-supervised learning method to characterize each plant of a crop field based on hidden relationships between vegetation indices. The vegetation indices are derived using hyperspectral data from around 400 to 1000 nm. We utilize a neural network embedding approach to create a mathematical space for different vegetation index bands (observed in plants of a crop field). The embedding space is contextual. That is if vegetation index band 1 and vegetation index band 2 appear in a part of tree 1, and vegetation index band 2 and vegetation index band 3 appear in a part of another tree, then all three vegetation index bands 1, 2, 3, will appear in the vicinity in the constructed embedding space. Therefore, a known correlation between one vegetation index and disease can be interpolated to another connected vegetation index. This will increase the potential for early detection of diseases and stresses.
 
The contribution of this paper is limited to the construction of the embedding space for vegetation index bands and plants. The paper evaluates the potential of the embedding space using machine learning applications. Analysis of connections with diseases and stresses remains a future direction of this work.

The contributions of this paper are summarized as follows:
\begin{itemize}
    \item We present a new self-supervised learning technique to create a distributed representation (contextual neural network embeddings) for vegetation index bands and trees using hyperspectral images. The method compresses high-dimensional hypespectral images of trees to low-dimensional embedding vectors.
    \item We introduce a unique way of segmenting trees and creating vegetation index bands to build effective inputs for self-supervised learning using a neural network. 
    \item We design new ways for experimentation on self-supervised learning with unlabeled hyperspectral imagery data in the agriculture sector.
\end{itemize}

\section{Related work}
\label{sec:related}
The use of hyperspectral imagery data is increasing in precision agriculture. Many crops are sensitive to stresses and express visible reflection of trauma or change through leaf colors. However, by the time the stress patterns become visible, harvest loss becomes inevitable in many cases. Researchers have on detection of different kinds of stresses -- such as insect and drought stresses  \cite{stress_1, stress_2} -- using hyperspectral images. Early detection of disease is another field of interest \cite{disease_1, disease_2} for agricultural and food scientists. Hyperspectral imagery data are used for quality and safety assessment of fruits \cite{quality_fruits} too. 

Estimation of different indicators to plant phenomena has also been a research focus. Kyaw et al. \cite{esti.photo_nitrogen} leveraged hyperspectral image data of leaves to estimate the photosynthetic capacity and nitrogen content. Li et al. \cite{esti.water} estimated leaf water content from similar datasets. 
Plant pathologists use different vegetation indices as indicators of plant health in many cases. Those indices could be utilized from the hyperspectral spectrum of plant or leaf-images \cite{veg_indices_review}. The use of vegetation index-based indicators is common in disease  \cite{veg_disease_1, veg_disease_2, veg_disease_4} and stress detection \cite{veg_stress_1, veg_stress_2}. Vegetation indices are also used for remote monitoring of plants \cite{veg_growth_1, veg_growth_2}. 
In our work, we create contextual embeddings for different vegetation index bands. The embeddings are likely to help scientists discover hidden patterns between disease and stress indicators.

In the agriculture domain, the use of hyperspectral images many times revolves around classification problems. For example, Hycza, Stere{\'n}czak, and Ba{\l}azy \cite{classi_poten} utilized hyperspectral images to classify tree species. Fricker et al. \cite{classi_convo_1},  Zhang, Zhao, and Zhang \cite{classi_convo_2}, and Miyoshi et al. \cite{classi_deep} also focused on plant species identification using classification algorithms. Different neural network models were applied for these classification problems. None of these classification models utilized distributed representations for better contextual classification, even though distributed representations using neural network embeddings have been proven effective in other domains, such as natural language processing \cite{word2vec_dist, word2vec_effi, bert, glove}.
In this research, we focus on creating a self-supervised neural network model for a distributed representation of vegetation index bands. We demonstrate the potential of such a distributed representation in several classification algorithms and contextual analysis of tress.


\section{Problem description}
\label{sec:problemdesc}
Let $T = \{t_1, t_2, t_3, ..., t_{|T|}\}$ be a collection of $|T|$ separate trees with hyperspectral pixels. For each hyperspectral pixel, a set of $|Indices|$ vegetation indices are computed, where $Indices = \{Index_1, \;Index_2, \newline \;Index_3, \;...,\; Index_{|Indices|}\}$. Each vegetation index, $Index_i$, is further divided into $|b|$ number of bands (ranges of values of the index), where the bands are $b_1, b_2, ..., b_{|b|})$. Therefore, there are $|b|\times|Indices|$ vegetation index bands. Let us represent the set of all vegetation index bands as $V$, where $V=\{Index_1^{b_1},\: Index_1^{b_2}, \: ..., \: Index_1^{b_{|b|}}, \newline Index_2^{b_1},\: Index_2^{b_2},\: ..., Index_2^{b_{|b|}},\: ..., Index_{|Indices|}^{b_{|b|}}\}=\{v_1, v_2, v_3, ..., \newline \; v_{|b|*|Indices|}\} $.
Each tree has $|S|$ segments of circular rings, where $S = \{s_1, s_2, s_3, ..., s_{|S|}\}$. $s1$ is in the center and is a circle. The rest of the segments are rings. Each ring and the center circle of a tree is called a context. The context set in the entire dataset is $C = \{c_{t_1}^{s_1}, c_{t_1}^{s_2}, ..., c_{t_1}^{s_{|S|}}, c_{t_2}^{s_1}, c_{t_2}^{s_2}, ..., c_{t_2}^{s_{|S|}}, ..., c_{t_{|T|}}^{s_{|S|}} \}.$ For $|T|$ trees, the total number of contexts is $|T|\times|S|$. 

This research aims to generate self-supervised models for all vegetation index bands and trees.
\begin{itemize}
    \item Create a self-supervised embedding model to provide vectors, $U=\{u_{Index_1}^{b_1},\: u_{Index_1}^{b_2},\: ..., u_{Index_1}^{b_{|b|}}, u_{Index_2}^{b_1},\: u_{Index_2}^{b_2},\: ..., u_{Index_2}^{b_{|b|}},\: ..., \newline u_{Index_{|Indices|}}^{b_{|b|}}\}=\{u_1, u_2, u_3, ..., u_{|b|*|Indices|}\}$, for all vegetation index bands.
    \item Generate embeddings for all trees $\tau=\tau_1, \tau_2, ... \tau_{|T|}$, where $\tau_i$ is composed of the embedding vectors of all $|S|$ segments of tree $t_i$. That is, the embeddings for tree $t_i$ is a vector of size $|S|\times|Indices|\times|u_i|$.
\end{itemize}

\section{Methodology}
\subsection{Data collection process}
The hyperspectral datasets were collected using an unmanned aircraft system (UAS) based on DJI Matrice 600 that was integrated with Resonon® Pika L hyperspectral push-broom imaging system (imaging spectroradiometer). The flights were conducted in cloudless weather conditions around solar noon over a Pecan and tornillo orchards near El Paso, Texas. The hyperspectral imager has a spectral range of around 400-1000nm, with about 300 spectral channels and a minimum spectral bandwidth of 2.1nm. The UAS was also equipped with a separate small-sized Real-Time Kinetic (RTK) Inertial Navigation System (INS) from SBG systems® for the imager. The UAS was flown at the height of about 100m above ground level with a 17mm fixed focus lens, felid-of-view (FOW) of 17.6 degrees. The data obtained were radiometrically corrected using the factory calibration data to compute radiance; radiance data were converted to surface reflectance (scaled to integer values between 0-10,000) via in-scene reference spectra from a calibration-tarp placed in the flight path, and finally, geo-rectified and resampled to produce a hypercube of 0.07m spatial resolution. The hypercube was used for further analysis and computation of indexes.

\subsection{Extracting trees}
\label{sec:extractTrees}
We ran the hyperspectral images through several prepossessing steps to extract the trees from other vegetation such as grass. The following subsections describe the steps.

Our observation is that pixels associated with leaves, the surrounding of a tree (such as the grass area), and tree shades can be characterized using the brightness of pixels. 
Let us assume that the brightness of pixel $P$ at wavelength $x$ is $P_x$. We found that most leaf pixels have $P_{780}>2500$, $P_{900}<6000$, and $P_{600}<1000$. Pixels in the surroundings and shades can be characterized the same way. However, some pixels in the surroundings with grass may have similar brightness to the leaves, while shades are easily distinguishable from the brightness property.

To ensure a better distinction between leaf pixels and surrounding pixels, we used two widely used vegetation indices for canopy or leaf area detection: (Eq. \ref{eq:ari2})  Anthocyanin Reflectance Index 2 (ARI2) \cite{veg_index}, and (Eq. \ref{eq:sipi}) Structure Intensive Pigment Index (SIPI) \cite{veg_index}. 

\begin{equation}
ARI2(P) = P_{800} * (\frac{1} {P_{550}} + \frac{1} {P_{700}})
\label{eq:ari2}
\end{equation}

\begin{equation}
SIPI(P) = \frac{P_{800} - P_{445}} {P_{800} + P_{680}}
\label{eq:sipi}
\end{equation}

\begin{figure}[t]
  \centering
  \includegraphics[width=126pt, height=224pt, angle =90]{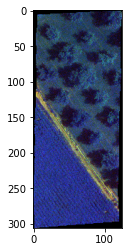}
  \centering
  \includegraphics[width=126pt, height=224pt, angle =90]{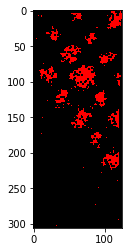}
  \caption{ (top) Before filtration, (bottom) After filtration.}
  \label{fig:beforeafter_fil}
\end{figure}

Given a pixel p, the following formula detects whether p is a leaf pixel (1) or not (0). 
\begin{equation}
\begin{aligned}
isLeaf(P) = 
\left\{
\begin{matrix} 1 & if ARI2(P) > \theta _{ARI2_{lf}} \: \& \: SIPI(P) > \theta _{SIPI_{lf}}\\
& \& \: P_{900} < 6000 \: \& \: P_{780} > 2500 \: \& \: P_{660} < 1000\\ 
0 & otherwise.
\end{matrix}\right.
\end{aligned}
\label{eq:filter_con}
\end{equation}

\noindent where $\theta _{ARI2_{lf}}=0.80$ and $\theta _{SIPI_{lf}}=0.88$. Both the thresholds $\theta _{ARI2_{lf}}$ and $\theta _{SIPI_{lf}}$ are empirically selected. Fig. \ref{fig:beforeafter_fil} (top) shows an image without filtration, and Fig. \ref{fig:beforeafter_fil} (bottom) shows the corresponding leaf pixels (in red) after the filtration of the non-leaf pixels using Eq. \ref{eq:filter_con}.

\subsubsection{Analyzing neighboring pixels to separate trees:}
The leaf pixels (Fig. \ref{fig:beforeafter_fil} (bottom)) are scattered points without any information on how they are connected within trees. To generate connectivity relationships between neighboring pixels, we divide an entire image to fit as many $k \times k$ non-overlapping grids as possible. We varied $k$ from 2 to 10, and k=4 resulted in the best outcome for most images. Let us say that the grids in an image are flattened as $G=\{g_1, g_2, ..., g_|G|\}$. Assume that $countlp(g_i)$ is the number of leaf pixels in grid $g_i$ of an image. 

\begin{equation}
isconnected(g_i)= \left\{\begin{matrix}
1 & if countlp(g_i) >= \theta_g\\ 
0 & otherwise
\end{matrix}\right.
\label{eq:comp_thresh}
\end{equation}

Here, $\theta_g$ the minimum number of pixels required to establish all the pixels inside the grid-connected. $\theta_g=10$ gave us reasonable connectivity between pixels of the same tree for $k=4$. That is, if a grid has 10 leaf pixels out of 16, then the grid is considered a container of connected pixels.

Consider that $g(p)$ is the grid where pixel $p$ exists. Non-leaf pixels within a grid that contains at least $\theta_g$ leaf pixels are marked for the detection of tree areas.

\begin{equation}
isInConnectedGrid(p)=\left\{\begin{matrix}
1 & if isconnected(g(p))=1\\ 
0 & otherwise
\end{matrix}\right.
\label{eq:isInConnectedGrid}
\end{equation}

\begin{figure}[t]
  \centering
  \includegraphics[width=126pt, height=224pt, angle =90]{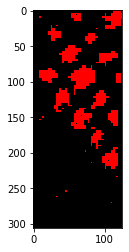}
  \Description{Connecting neighboring pixels}
  \caption{Connecting neighboring pixels.}
  \label{figure:conn_neigh_pix}
\end{figure}

Fig. \ref{figure:conn_neigh_pix} shows the pixels that are detected as within connected grids (using Eq. \ref{eq:isInConnectedGrid}) for our running example.

\subsubsection{Indexing the trees:}
\begin{figure}[t]
  \centering
  \includegraphics[width=126pt, height=224pt, angle =90]{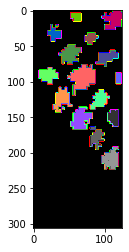}
  \Description{Numbering the trees}
  \caption{Indexing the trees.}
  \label{fig:num_tree}
\end{figure}

For pixel $p$, for which $isInConnectedGrid(p)$ $=1$, we apply depth first search (DFS) to include all the leaf pixels of a tree. When DFS is exhausted for one  pixel, then the algorithm moves to an unexplored pixel with $isInConnectedGrid(p)=1$. The algorithm continues till there is no more pixel to traverse. A constructed tree with less than 40 pixels is not actually a tree, based on our observation, and hence removed. Also, we only keep leaf pixels within a tree area. We index the trees, $T = \{t_1, t_2, ..., t_{|T|}\}$, based on the sequence DFS created them. 
Fig. \ref{fig:num_tree} shows the pixels of the trees.

\subsection{Computing pixel-wise vegetation indices}
Vegetation indices play a vital role in analyzing the health of trees. For instance, anthocyanins are responsible for colors in fruits and vegetables. The level of anthocyanins can be computed from a pixel using Anth ocyanin Reflectance Index 1(\textit{ARI1}) and Anthocyanin Reflectance Index 2(\textit{ARI2}). The formulas are provided in Table \ref{tab:veg_ind}. We calculated 21 hyperspectral vegetation indices, which are listed in Table \ref{tab:veg_ind}. For some of the vegetation indices, not all wavelengths were available in the hyperspectral images. In such cases, we used the closest available wavelength to calculate the vegetation index. For example, according to the Eq. \ref{eq:sipi} (Structure Insensitive Pigment Index - SIPI), we need $P_{445}$, $P_{680}$, $P_{800}$. The exact wavelengths of 680 and 800 were not available in the data. We used 680.47 nm instead of 680 because 680.47 nm was the closest to 680. Similarly, we used 799.94 nm instead of 800.

\begin{table*}
\begin{center}
\caption{Hyperspectral vegetation indices}
\begin{tabular}{ |l|l|l| } 
 \hline
 \textbf{SL.} & \textbf{Hyperspectral Vegetation Index} & \textbf{Equation}\\
 \hline
 01 & Anthocyanin Reflectance Index 1 & $ARI1=\frac{1}{p_{550}} - \frac{1}{p_{700}}$\\ 
 \hline
 02 & Anthocyanin Reflectance Index 2 & $ARI2=p_{800}\left (\frac{1}{p_{550}} - \frac{1}{p_{700}}  \right )$\\ 
 \hline
 03 & Atmospherically Resistant Vegetation Index & $ARVI=\frac{NIR-(Red - \gamma (Blue - Red))}{NIR+(Red - \gamma (Blue - Red))}$\\
 \hline
 04 & Carotenoid Reflectance Index 1 & $CRI1=\frac{1}{p_{510}} - \frac{1}{p_{550}}$\\ 
 \hline
 05 & Carotenoid Reflectance Index 2 & $CRI2=\frac{1}{p_{510}} - \frac{1}{p_{700}}$\\ 
 \hline
 06 & Enhanced Vegetation Index & EVI=$\frac{NIR-Red}{NIR+6.0\cdot Red - 7.5 \cdot Blue + 1}$\\
 \hline
 07 & Modified Chlorophyll Absorption Reflectance Index & $MCARI=p_{700} - p_{670} - 0.2 (p_{700} - p_{550}) \left (\frac{p_{700}}{p_{670}}\right)$\\ 
 \hline
 08 & Modified Chlorophyll Absorption Reflectance Index Improved & $MCARI2= \frac{ 1.5 \cdot (2.5 \cdot (NIR-Red) - 1.3 \cdot (NIR - Green)) } {\sqrt{(2\cdot NIR + 1)^2 - (6 \cdot NIR - 5\cdot\sqrt {Red} - 0.5}}$\\ 
 \hline
 09 & Modified Red Edge Normalized Vegetation Index & $MRENVI=\frac{p_{700} - p_{705}}{p_{750} + p_{705} - 2 \cdot p_{445}}$\\
 \hline
 10 & Modified Red Edge Simple Ratio Index & $MRESRI=\frac{p_{750}-p_{445}}{p_{705}-p_{445}}$\\ 
 \hline
 11 & Normalized Difference Vegetation Index & $NDVI=\frac{NIR-RED}{NIR+RED}$\\ 
 \hline
 12 & Photochemical Reflectance Index & $PRI=\frac{p_{531}-p_{570}}{p_{531}+p_{570}}$\\
 \hline
 13 & Plant Senescence Reflectance Index & PSRI=$\frac{p_{680}-p_{500}}{p_{750}}$\\
 \hline
 14 & Red Edge Normalized Difference Vegetation Index & $RENDVI=\frac{p_{750}-p_{705}}{p_{750}+p_{705}}$\\
 \hline
 15 & Simple Ratio Index & $SRI=\frac{NIR}{Red}$\\ 
 \hline
 16 & Structure Insensitive Pigment Index & $SIPI=\frac{p_{800}-p_{445}}{p_{800}+p_{680}}$\\
 \hline
 17 & Transformed Chlorophyll Absorption Reflectance Index & $TCARI=3 \left [p_{700} - p_{670} - 0.2 (p_{700} - p{550}) \left (\frac{p_{700}}{p_{670}}  \right )  \right ]$\\ 
 \hline
 18 & Vogelmann Red Edge Index 1 & $VREI1=\frac{p_{740}}{p_{720}}$\\ 
 \hline
 19 & Vogelmann Red Edge Index 2 & $VREI2= \frac{p_{734}-p_{747}}{p_{715}-p_{726}}$\\
 \hline
 20 & Vogelmann Red Edge Index 3 & $VREI3=\frac{p_{734}-p_{747}}{p_{715}+p_{720}}$\\ 
 \hline
 21 & Water Band Index & $WBI=\frac{p_{970}}{p_{900}}$\\ 
 \hline
\end{tabular}
\label{tab:veg_ind}
\end{center}
\end{table*}

\subsection{Analyzing the relationship between segments of a tree and vegetation indices}
\label{sec:relbtwsegandvind}

\begin{wrapfigure}{r}{0.18\textwidth}
  \includegraphics[width=90pt, height=90pt]{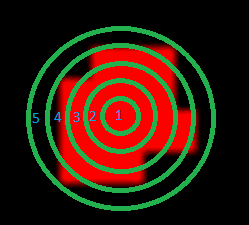}
  \caption{Segments of a tree.}
  \label{fig:seg}
\end{wrapfigure}

Our observation is that, in a drone-view of a tree, most vegetation indices vary between the central part of a tree and the outer part. We divide each tree area into $S = \{s_1, s_2, s_3, ..., s_{|S|}\}$ circular segments. We consider the center of the line drawn between the  two furthest pixels of a tree as the center of the tree. The radius of the tree is considered to be half of the length of the line between the two furthest points of the tree. The tree area is divided into $|S|$ parts. The first segment, $s_1$ is the circle from the center with a radius of $\frac{radius}{|S|}$. Segment $s_2$ is the ring between the radius $\frac{radius}{|S|}$ to $\frac{2*radius}{|S|}$, so and so forth. Then, we compute the average of each vegetation index of all  the pixels in each segment. Figure \ref{fig:seg} shows a sample of the segments of a tree area, where the color red represents the tree area. The green circles separate the segments. 

For every vegetation index, we checked if the vegetation index keeps increasing or decreasing in at least $l$ segments from the center to the outer part, where $l\leq|S|$.
 
Consider $v_{\text{avg}}(s_i)$ the average of a vegetation index of the pixels in segment $s_i$. To check if there are $l$ segments with increasing or decreasing values of a vegetation index in the segments, we use the following formula.

\begin{equation}
pc(l) = 
\left\{\begin{matrix}
l & if \: max\{\sum_{i=1}^{|S|} v_{\text{avg}}(s_i) > v_{\text{avg}}(s_{i-1}), \\
 & \sum_{i=1}^{|S|} v_{\text{avg}}(s_i) < v_{\text{avg}}(s_{i-1})\} = l\\ 
0 & otherwise
\end{matrix}\right.
\label{eq:check_seg_len}
\end{equation}

\noindent The formula returns $l$ if there are $l$ such segments with increasing or decreasing average values of the vegetation index. Otherwise, it returns 0.

The following formula gives the largest number of segments in increasing or decreasing values for a given vegetation index. 

\begin{equation}
    pl_{max} = \max_{l = 2}^{|S|} \: pc(l)
\label{eq:max_seg_len}
\end{equation}

\begin{figure}[b]
  \centering
  \includegraphics[width=\linewidth]{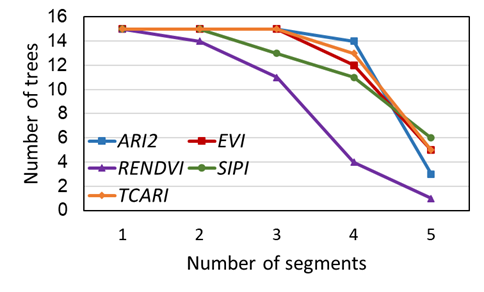}
  \caption{Relationship between the number of segments with increasing or decreasing vegetation index. Plots are given for five vegetation indices in five lines.}
  \label{fig:seg_wis_tree_area}
\end{figure}

Fig. \ref{fig:seg_wis_tree_area} shows the relationship between the number of segments and the number of trees that have increasing or decreasing vegetation index within the segments. The experiment was done with 15 trees collected from the same hyperspectral image. The figure shows the number of trees having a vegetation index in $l$ segments where $l$ varies from 1 to 5. Each line represents a vegetation index. While most vegetation indices follow an increasing or decreasing pattern in most trees, some do not have any increasing or decreasing patterns. For example, RENDVI (Red Edge Normalized Difference Vegetation Index) does not have increasing or decreasing patterns within the segments of the trees, as observed in Fig. \ref{fig:seg_wis_tree_area}. Note that there are only four trees (out of 15) with increasing or decreasing patterns with the number of segments=4 for RENDVI. The corresponding line represents that RENDVI does not have the center of the tree different from the outer parts.

\begin{figure*}[t]
  \centering
  \includegraphics[width=\linewidth]{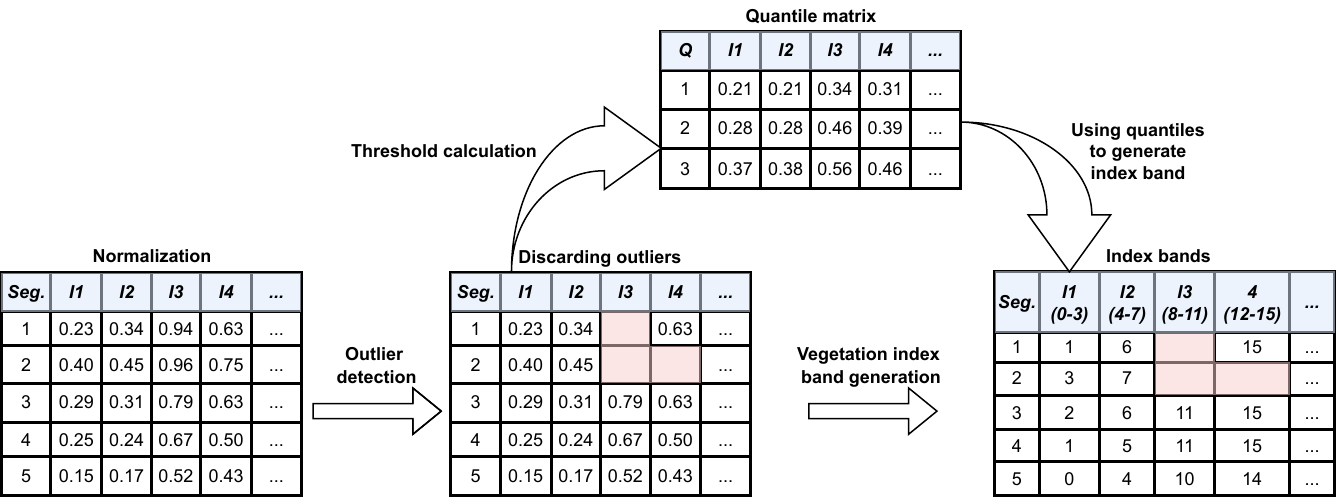}
  \caption{How the tree data are prepared for an input of embedding models. $I_i$ refers to vegetation index $Index_i$.}
  \label{fig:pre}
\end{figure*}

Not all the vegetation indices are plotted in Fig.\ref{fig:seg_wis_tree_area} to ensure picture clarity. Our observation is that most vegetation indices have increasing or decreasing values from the center toward the outer region, making the choice of creating segments within a tree for feature generation a logical one. With more than five segments, we observed that the number of pixels is not workable. With a lesser number of segments, the increasing or decreasing patterns are not captured well. Therefore, in our experiments, we divided each tree into five segments.

\subsection{Input for the embedding model}
As illustrated in Fig. \ref{fig:pre}, how tree segments go through a series of changes to create input data for an embedding model. Each segment of a tree has an average of 21 vegetation indices. The first step is to normalize each vegetation index based on all segments of all trees using min-max normalization. That brings all indices within the range of 0 to 1. 

After the normalization, we detect outlier values for each vegetation index. For each vegetation index, outliers are detected for all values in all segments of all trees. We use a box-plot method to detect outliers. $Q1_{Index_j}$ is the first quartile, $Q2_{Index_j}$ is the median, and $Q3_{Index_j}$ is the third quartile of the vegetation index $Index_j$. Then, the interquartile range is given by the following formula.

\begin{equation}
    IQR_{Index_j} = Q3_{Index_j} - Q1_{Index_j}
\label{eq:out_iqr}
\end{equation}

\noindent whereas, the lower-bound is 
\begin{equation}
    \text{lb}_{Index_j} = Q1_{Index_j} - 1.5*IQR_{Index_j}
\label{eq:out_min}
\end{equation}

\noindent and the upper-bound is
\begin{equation}
    \text{ub}_{Index_j} = Q3_{Index_j} + 1.5*IQR_{Index_j}
\label{eq:out_max}
\end{equation}

A value of $Index_j$ in segment $s_i$ of a tree $t_k$ smaller than the lower-bound or greater than the upper-bound is marked as an outlier, with a negative value. 
\begin{equation}
t_k^{s_i, Index_j} = \left\{\begin{matrix}
t_k^{s_i, Index_j} & if \: \text{lb}_{Index_j} <=t_k^{s_i, Index_j} <=  \text{ub}_{Index_j}\\ 
-1000000 & otherwise
\end{matrix}\right.
\label{eq:out_check}
\end{equation}

\noindent In Fig. \ref{fig:pre}, the outlier cells after detection are marked with a light shade of red. After the removal of outliers, we create vegetation index bands. Each vegetation index range is divided into four bands in such a way that each band has an equal number of observations in all segments of all trees. We achieve such equidistribution by creating three thresholds at quartile 1, quartile 2, and quartile 3 of the outlier-discarded data. The thresholds to create the bands are shown in the quartile matrix of Fig. \ref{fig:pre}.

Now, each vegetation index in every segment of a tree is converted to an integer representing which band the index falls into, as shown in the right-most matrix of Fig. \ref{fig:pre}. The number of vegetation index bands $|V|$ becomes $4\times21 = 84$.

\subsection{Building a self-supervised model to generate embeddings for vegetation index bands}
As stated in Section \ref{sec:problemdesc}, one of the objectives is to create a self-supervised embedding model $U$ for each vegetation index band: $U=\{u_{Index_1}^{b_1},\: u_{Index_1}^{b_2},\: ..., u_{Index_1}^{b_{|b|}}, u_{Index_2}^{b_1},\: u_{Index_2}^{b_2},\: ..., u_{Index_2}^{b_{|b|}},\: ..., \newline u_{Index_{|Indices|}}^{b_{|b|}}\}=\{u_1, u_2, u_3, ..., u_{|b|*|Indices|}\}$. The current state of the art embedding models are Word2vec \cite{word2vec_effi, word2vec_dist}, Glove \cite{glove}, BERT \cite{bert}, which are suitable for text data. These models leverage semantic relationships of words within a context window. Such semantic relationships do not exist in our problem with vegetation indices using hyperspectral imagery. However, as described in Section \ref{sec:relbtwsegandvind}, trees have increasing or decreasing order of vegetation indices from the center to outer segments, making each segment a potential context for vegetation index bands. In such a unique scenario of vegetation index for trees, we have $|V|=84$ bands which is analogous to the vocabulary in natural language terminology.

\subsubsection{A Word2vec-inspired model:}
We constructed two Word2vec models, one using the Continuous Bag of Words (CBOW) approach and the other using the Skip-graph model, that is commonly used in generating embedding vectors in natural language processing. With our data, each band was a word (with a vocabulary size of 84), each tree is a document, and each segment of a tree is considered a sentence/context. Neither of the models was able to produce contextual embeddings for the vegetation index-based data generated from high-spectral images. One of the reasons is that Word2vec captures the semantic relationships from a large number of samples (documents). Even though our data are high-spectral (high-dimensional), the number of sentences or segments in the trees is limited to generate a solid neural network embedding-based mathematical space.

The results gained from the word2vec inspired us to construct a neural network model that performs well with a limited number of input samples, as described in the following subsection.

\subsubsection{Embedding model to generate embeddings for vegetation index bands:}
\label{sec:embeddingmondel}
The objective of creating embeddings is to bring two contextually similar vegetation index bands close by in a mathematical space. If two vegetation index bands are found in the same tree segments, then these two vegetation index bands must be contextually similar. Assume that two vegetation index bands $v_i$ and $v_j$ are contextually similar. Another pair of vegetation index bands, $v_j$ and $v_k$, are also contextually similar. Even though $v_i$ and $v_k$ bands never appeared together in any tree segment, they should be contextually connected via $v_j$. Regular vector space, such as direct use of vegetation index bands or hyperspectral ranges, will not provide such contextual similarity. With an embeddings space that correctly puts the vegetation index bands in the mathematical space,  $v_i$, $v_j$, and $v_k$ will be in the same neighborhood indicating contextual proximity. 

To encode the appearance of a pair of bands, $v_i$ and $v_j$, we use the Jaccard similarity between the segments where they appear. Let $s(v_i)$ be the set of segments where the vegetation index band $v_i$ was observed. Then the Jaccard similarity between the existence of $v_i$ and $v_j$ is:

\begin{equation}
J(v_i,v_j) = 
\frac{|s(v_i) \cap s(v_j)|}
{|s(v_i) \cup s(v_j)|}
\label{eq:jac_sim}
\end{equation}

\begin{figure}[t]
  \centering
  \includegraphics[width=200pt]{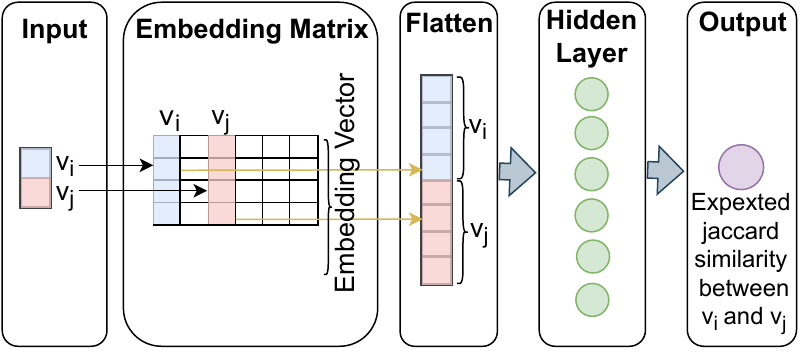}
  \caption{Neural network architecture to generate embeddings using hyperspectral tree imagery and vegetation index data.}
  \label{fig:modelarch}
\end{figure}

Fig. \ref{fig:modelarch} shows the neural network model to generate the embeddings using the Jaccard similarity between all $v_i$ and $v_j$ pairs of embedding bands in all trees as training information. The input to the neural network is two band IDs, where IDs range between 1 and the number of vegetation index $|V|$. The forward propagation moves via an embedding layer. The embedding matrix has $|V|$ columns and 64 rows, generating an embedding vector of length 64 for each vegetation index. For the $v_i$ and $v_j$ inputs, the $i$th and $j$th columns of the embedding matrix are flattened to construct a column vector $u_{ij}$ of length 128. Forwarding from that, via a hidden layer, the loss for one pair of vegetation indices can be computed as:

\begin{equation}
Loss (i, j) = ((w_h)^T u_{ij})^T w_o- J(v_i,v_j)
\label{eq:loss_ij}
\end{equation}

$w_h$ is the weight matrix used between the flattened embedding vector $u_{ij}$ and the hidden layer. $w_o$ is the weight vector between the hidden layer and the output. That is, $w_o$ has a length equal to the number of hidden nodes.

The adam optimizer minimizes the following loss function.

\begin{equation}
FinalLoss = \frac{1}{n} \sum_{i=1}^{|V|}\sum_{j=i+1}^{|V|} \left( Loss(i, j) \right)^2
\label{eq:loss_ij}
\end{equation}

\noindent where $n=\sum_{i=1}^{|V|}\sum_{j=i+1}^{|V|} 1$. We used leaky relu as the activation function. The hidden layer had 600 nodes, with a 20\% drop-out rate.

\subsection{Constructing tree vectors using vegetation index band embeddings}
\label{sec:constrees}
As explained in Section \ref{sec:embeddingmondel}, each vegetation index band $v_i$ is represented as an embedding vector $u_i$ of length 64. Each tree has $|S|$ segments, which is 5 in our case, as explained in Section \ref{sec:relbtwsegandvind}. Each segment has $|Indices|=21$ vegetation indices. To construct a tree from the vegetation index bands, we combine the embeddings of the vegetation index bands of each tree segment in a sequence. The segments appear from the center to the outer part of the tree. Hence, a tree $t_i$ is represented by a vector $\tau_i$ of length $|S|\times|Indices|\times|u_i|=5\times21\times64=6720$. 

As shown in Figure \ref{fig:pre}, the tree representation might have some vegetation index bands detected as outliers. An embedding vector of a vegetation index $v_i$ for an outlier in a segment $s_p$ is copied from the vegetation index band embedding $u_i$ of a non-outlier vegetation index band of another nearest segment $s_q$. If an outlier vegetation index band in a segment $s_p$ has two nearest segments $s_{p-k}$ and $s_{p+k}$, where $k$ is either 1 or 2, the average of two $u_i$'s of those two segments, $s_{p-k}$ and $s_{p+k}$, will become the vegetation index band $u_i$ of the outlier segment $s_p$.

\section{Experimental results}
To evaluate the potential of the embeddings generated for vegetation index bands and constructed for trees, we use two hyperspectral image datasets of trees. The first dataset has 81 pecan trees. The second dataset contains 198 tornillo trees. For each dataset, we extracted the trees from large aerial hyperspectral images of the crop fields using the method described in Section \ref{sec:extractTrees}.

In this section, we seek to answer the following questions.
\begin{enumerate}
    \item Does the embedding space constructed for the vegetation index bands have structures capturing relationships between them? (Section \ref{sec:exp:tsne})

    \item How are embedding-based clusters of trees different from direct vegetation index-based clusters? (Section \ref{sec:exp:purity})

    \item How well does the embedding-based tree representation perform in creating discriminative features compared to a direct vegetation index-based representation of trees? (Section \ref{sec:exp:clustclassification})

    \item How can we characterize groups of trees constructed from an embedding-based representation? (Section \ref{sec:exp:centroid})
    
    \item How can we retrieve the context of vegetation index bands using an embedding-based representation? (Section \ref{sec:exp:contextual})
\end{enumerate}

\subsection{Does the embedding space of vegetation index bands capture relationships between vegetation indices?}
\label{sec:exp:tsne}
\begin{figure}[t]
  \centering
  \includegraphics[width=\linewidth]{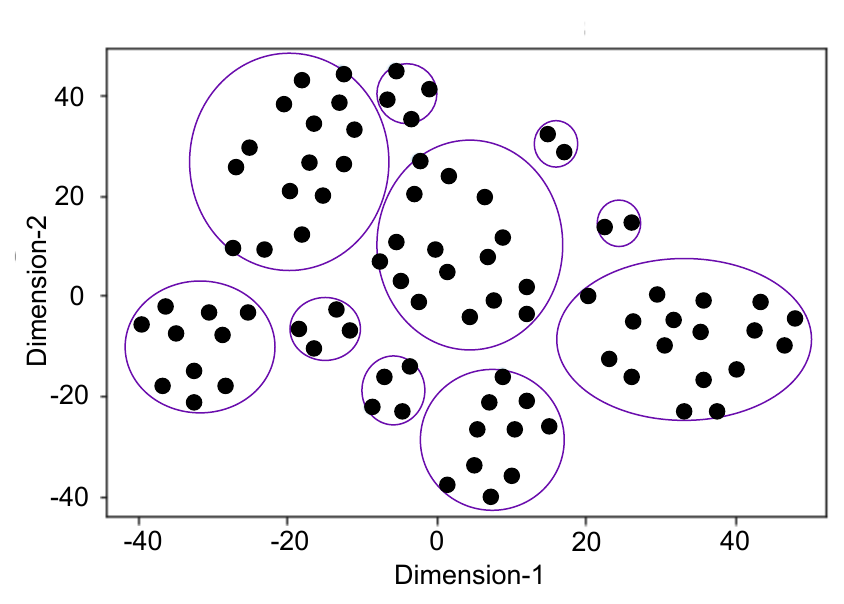}
  \caption{A two-dimensional t-SNE Plot for the embedding vectors of the vegetation index bands. There are $21\times4=84$ points in the plot for 21 vegetation indices.}
  \label{fig:tsne}
\end{figure}

Using a t-Distributed Stochastic Neighbor Embedding (t-SNE) plot of two dimensions in Fig. \ref{fig:tsne} (constructed from the 64-dimensional embedding space), we observe that embedding vectors of the vegetation index bands form some structures. Each point in the figure represents the embedding of a vegetation index band. Some groups are manually drawn in Fig.\ref{fig:tsne}. The plot only illustrates that there are structures in the embedding space, but it does not demonstrate the potential of using the embeddings for analyzing trees.

The following subsections provide experiments demonstrating the potential of the use of embeddings of vegetation index bands in analyzing hyperspectral imagery of plants.

\subsection{Difference between embedding-based and vegetation index-based clusters}
\label{sec:exp:purity}
Even before we study the potential of embeddings, a crucial question is -- if trees represented by embeddings provide a different insight than the same trees represented by direct vegetation indices in segments of the trees. Each tree vector in our embedding-based representation has a length of $6720$. For the direct vegetation index-based representation, each tree is represented as a vector of length $|S|\times|Indices|=5\times21=105$.
 
\begin{table}[b]
\begin{center}
\caption{Confusion matrix using embedding-based clusters and vegetation index-based clusters of trees for the Pecan dataset.}
\begin{tabular}{ |c|c|c|c|c| } 
 \hline
 \textbf{\pbox{55pt}{$\rightarrow\text{Direct}$\newline$\downarrow\text{Embedding}$}} & \textbf{Cluster 1} & \textbf{Cluster 2} &\textbf{Cluster 3} & \textbf{Cluster 4}\\
 \hline
 \textbf{Cluster 1} & 0 & 2 & 15 & 1\\
 \hline
 \textbf{Cluster 2} & 19 & 1 & 0 & 0\\
 \hline
 \textbf{Cluster 3} & 3 & 19 & 2 & 0\\
 \hline
 \textbf{Cluster 4} & 0 & 3 & 1 & 15\\
 \hline
\end{tabular}
\label{tab:confusionmat}
\end{center}
\end{table} 
 
For each representation, we apply k-means with k=4. Then we constructed a confusion matrix illustrating the size of intersections of trees between cluster $i$ of embedding-based clustering and cluster $j$ of direct vegetation index-based clustering. $i$ and $j$ vary between 1 to 4 because the number of clusters is 4. As observed in the confusion matrix of Table \ref{tab:confusionmat}, none of the embedding-based clusters solely intersect with a direct vegetation index-based cluster. If each embedding-based cluster had a sole mapping with one direct vegetation index-based cluster, then exactly four cells of the confusion matrix would have non-zero values.

The pecan dataset was used for Table \ref{tab:confusionmat}. We use purity to compute a scalar value representing overlaps between the clusters of two clusterings.
\begin{equation}
purity =  \frac{\sum_{i = 1}^{k} \max_{j = 1}^{k} (g_{i} \cap h_{j})} {|T|}
\end{equation}

\noindent where $k$ is the number of clusters in each clustering and $g_i$ refers to the $i$th cluster of the embedding-based clustering, and $h_j$ refers to the $j$th cluster of the direct vegetation index-based clustering. 

For Table \ref{tab:confusionmat}, the computed purity is 0.84, which indicates that around 84\% of the trees agree on the clustering results using embedding and direct vegetation index-based approach. The summary of this experiment is that the embedding-based approach does not result in the same clustering result as the direct vegetation-based approach. We ran a similar test with the tornillo dataset, which also resulted in the same conclusion that the embedding-based approach gives a different clustering result. 

While this subsection focuses on the uniqueness of the embedding-based approach compared to the direct use of vegetation indices, the following subsections provide experiments on the quality of the embedding vectors.

\subsection{A comparison between  embedding-based and direct vegetation index-based representations}
\label{sec:exp:clustclassification}

Section \ref{sec:constrees} explains how we constructed tree vectors using embeddings of the vegetation index bands. To evaluate the potential of the constructed embeddings of the trees, we compare the performance of a classification application between our embedding-based tree representation and a tree representation that directly uses vegetation indices of each segment of a tree. 

An issue in framing a classification problem is that we need labeled data. Since the dataset is newly created, there are no available classification labels. To mitigate the issue of labeling the data, we created four clusters of the available trees using the k-means clustering algorithm. Then we used the outcome of the k-means clustering as class labels for the trees. For a dataset, we created two clusterings of trees, one with embedding-based features and the other using direct vegetation index-based representation. For a fair comparison, we used embedding-based clusters of trees as labels when we used classification algorithms for embedding-based trees. We used direct vegetation index-based clusters when we used classification algorithms for direct vegetation index-based trees. We used both the pecan and tornillo datasets in this study.

\begin{figure}[t]
\centering
\begin{tabular}{@{}c@{}c} 
\includegraphics[width=120pt]{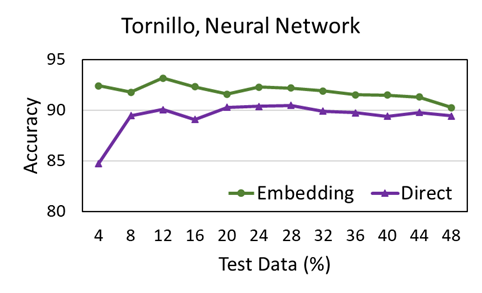}&\includegraphics[width=120pt]{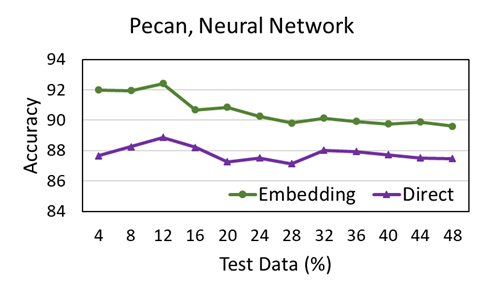}\\
\includegraphics[width=120pt]{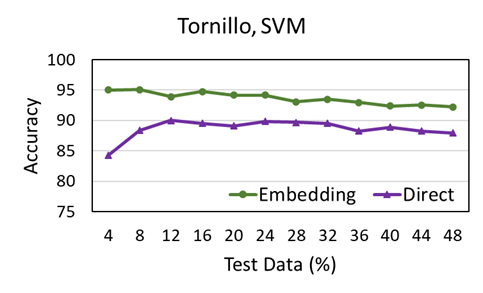}&\includegraphics[width=120pt]{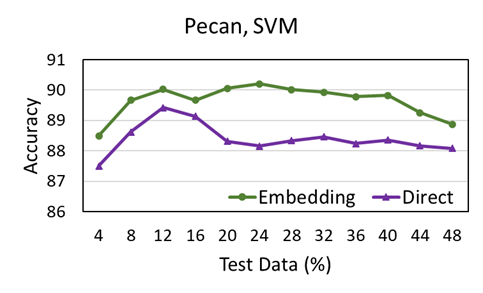}\\
\includegraphics[width=120pt]{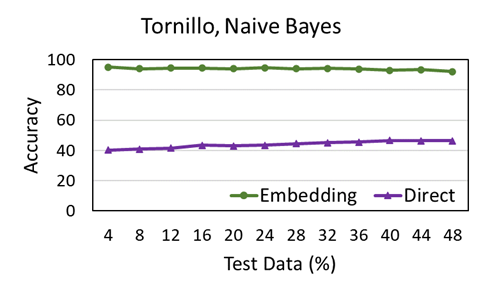}&\includegraphics[width=120pt]{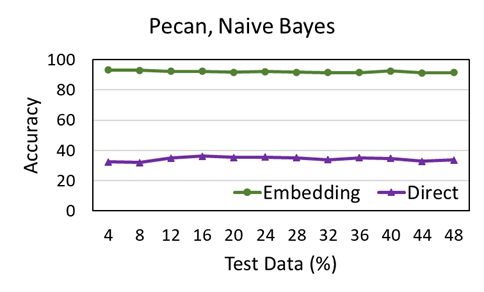}\\
\includegraphics[width=120pt]{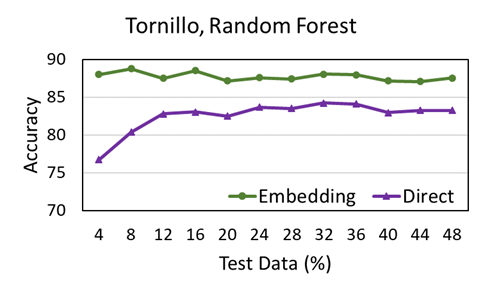}&\includegraphics[width=120pt]{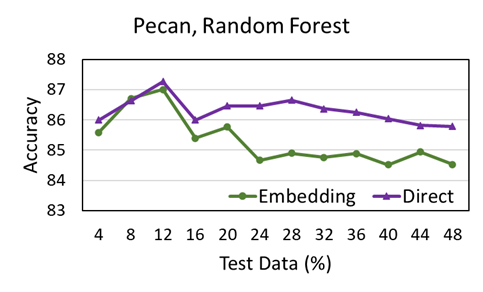}\\
\includegraphics[width=120pt]{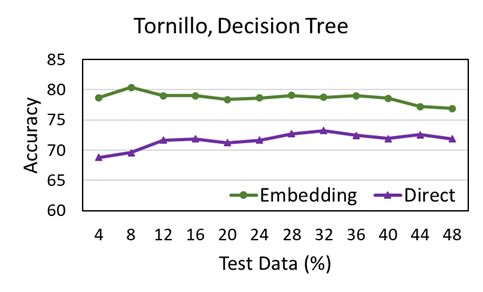}&\includegraphics[width=120pt]{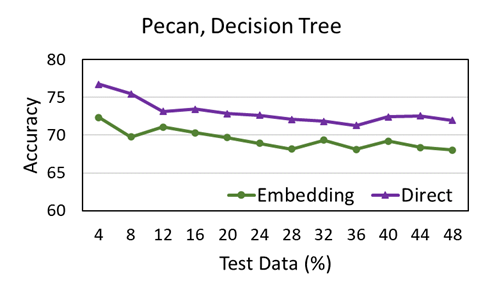}\\
\end{tabular}
\caption{Classification accuracy comparison between our embedding-based approach and direct veg. index-based approach. (left) Tornillo dataset and (right) Pecan dataset.}
\label{fig:eva_tree_clus}
\end{figure}

We applied five different classification algorithms, for each of the two representations (embedding-based and direct vegetation index-based), for each of the two datasets (pecan and tornillo). The five classification algorithms are neural network, support vector machine, naive Bayes, random forest, and decision tree. Fig. \ref{fig:eva_tree_clus} demonstrates the results. For each dataset and each algorithm, we performed the experiments with an increasing test data percentage of  4\%, 8\%, 12\%, 16\%, .., 48\% (that is, decreasing training portion). A repetition of 100 different random splits was taken at each percentage. The reported accuracy is the average of 100 computed accuracies of predicting the cluster labels of the test data. The five plots on the left are for tornillo, and the five plots on the right are for pecan plants.

As observed in the left-side plots of Fig. \ref{fig:eva_tree_clus}, our embedding-based representation results in better accuracies compared to the direct vegetation index-based representation, using all the five algorithms. On the right side, we observe that our embedding-based approach performs better for all splits using neural network, support vector machine, and naive Bayes classification. However, the direct vegetation index-based classification performs better for random forest and decision tree-based classification. The explanation is that the mechanics of the decision tree and random forest work in such a way that the best discriminating feature from the feature vector is selected to make a decision. The number of direct vegetation index features (105) is lesser than the number of embedding features 6720, making a single vegetation index feature more powerful than a single embedding-based feature. Moreover, embedding features are powerful and contextual when all of the features are used, not just one or a subset of the features.

Even though with a larger dataset (e.g., tornillo), our embedding-based approach demonstrates better classification results with all the five classification algorithms, we advocate the use of embedding-based approaches in classification techniques where all features are used in all aspects of an algorithm, such as neural network, support vector machine, and naive Bayes. The mechanics of decision trees and random forests do not support the actual use of embedding-based features.

\subsection{Characterizing clusters of trees retrieved from the embedding-based clustering}
\label{sec:exp:centroid}
As demonstrated in Section \ref{sec:exp:clustclassification}, the embedding-based tree representations have more contextual ability for classification. This indicates that the contextual features can be used for studying trees further. A complexity is that the features of embedding-based tree representations are not understandable for human interpretation. However, it is possible to characterize each embedding-based cluster based on the original vegetation index-based features of the trees.

To characterize each cluster, we take the vegetation index representation of all the trees of an embedding-based cluster and compute a vegetation index-based centroid:

\begin{equation}
\text{centroid}_\text{veg.Ind}(g_i) =
\frac
 {\sum_{j = 1}^{|g_i|}{( v(t_j))}}
 {|g_i|}
\end{equation}

\noindent where $g_i$ is the $i$th embedding-based cluster and $v(t)$ gives the direct vegetation index-based vector of tree $t$.

\begin{table}[t]
\begin{center}
\caption{Embedding cluster characterization by vegetation index bands.}
\label{tab:closestvegind}
\begin{tabular}{ |@{}c@{}|@{}c@{}|@{}c@{}|@{}c@{}|@{}c@{}|@{}c@{}| } 
\hline
\textbf{Clusters} & \textbf{1} & \textbf{2} &\textbf{3} & \textbf{4} & \textbf{5}\\
\hline
Cluster1 & Low SRI & Low ARVI & Low SIPI & Low NDVI & Low CRI1\\
\hline
Cluster2 & Mid TCARI & Mid MCARI & Mid ARI2 & Mid ARI1 & High ARI1\\
\hline
Cluster3 & Mid WBI & Low SRI & Mid ARI2 & Low ARI2 & Mid ARI1\\
\hline
Cluster4 & Low CRI1 & Low CRI2 & Low VREI1 & Low SRI & Low SIPI\\
\hline
\end{tabular}
\end{center}
\end{table}

Given an embedding-based cluster, $g_i$, we can now characterize the cluster by some top vegetation indices using the ascending order of the following computed values for the vegetation indices:

\begin{equation}
{d}_{g_i} =  
\frac
{\sum_{j = 1}^{|g_c|}( v(t_j)-\text{centroid}_\text{veg.Ind}(g_i)
)^2}
{|g_i|}
\end{equation}

\noindent ${d}_{g_i}$ is a vector of length 105 and represents the difference of each of the 105 vegetation index bands with the center of the $i$th embedding-based cluster. That is, each embedding-based cluster can now be characterized using these vegetation index bands. For example, Table \ref{tab:closestvegind} shows the top five vegetation index bands for four embedding-based clusters. Cluster 1 has low values of Simple Ratio
Index (SRI), low Atmospherically Resistant
Vegetation Index (ARVI), so and so forth. Cluster 2 exhibits mid-level bands of Transformed Chlorophyll
Absorption Reflectance Index (TCARI) and mid-level values of Modified Chlorophyll Absorption Reflectance Index (MCARI). Similarly, Clusters 3 and 4 have characterizations using the vegetation index bands. Such characterizations of clusters help scientists in plant pathology study indications of stressors and diseases in a group of plants.

\subsection{Contextual analysis of veg. index bands}
\label{sec:exp:contextual}
A powerful aspect of creating embeddings for vegetation index bands is that scientists can study contextual vegetation indices for each crop field. In Table \ref{tab:contextualNN}, we provide three nearest neighbors of several vegetation index bands both using the embedding space and directly using the vegetation index bands. For the embeddings, nearest neighbors were computed based on the distance between each pair of embedding vectors. For the direct vegetation index bands, nearest neighbors were computed based on the Jaccard similarity of two binary vectors representing the appearance of two vegetation index bands in tree segments.  

Table \ref{tab:contextualNN}(a) shows that the contextual (embedding-based) second nearest neighbor of low EVI (Enhanced Vegetation Index) is low VREI1 (Vogelmann Red Edge Index 1). In contrast, Table \ref{tab:contextualNN}(b) demonstrates that the second nearest neighbor of low EVI is low NDVI (Normalized Difference Vegetation Index). Other rows of Table \ref{tab:contextualNN}(a) and (b) also reflect differences. Embedding-based approaches retrieve hidden connections in the feature space, whereas direct approaches look for the direct similarity of the vectors.

Mid SRI's contextual nearest neighbor being mid SIPI does not necessarily mean that plants exhibit mid SRI and mid SIPI together. Rather, it means that they are contextually connected. A contextual connection may be established because mid SRI and mid ARVI appear in Tree 1, and mid ARVI and mid SIPI appear in Tree 2. Therefore, mid SRI $\rightarrow$ mid ARVI $\rightarrow$ mid SIPI could be the reason why mid SRI and mid SIPI are closer in the embedding space. No such contextual relationships are observed in a direct vegetation index-based space.

\begin{table}[t]
\begin{center}
\caption{Nearest vegetation index of a given one. V. High indicates Very High.}
\label{tab:contextualNN}
(a) Nearest vegetation index bands using embedding space.
\begin{tabular}{|@{}c@{}|@{}c@{}|@{}c@{}|@{}c@{}|} 
 \hline
 \textbf{Veg. Ind. band} & \textbf{1} & \textbf{2} &\textbf{3}\\
 \hline
 Low EVI & V. High PSRI & Low VREI1 & Low MCARI2\\
 \hline
 V. High MRENVI & V. High MRESRI & V. High RENDVI & Low VREI2\\
 \hline
 V. High PSRI & Low EVI & Low MCARI2 & High WBI\\
 \hline
 Mid SRI & Mid SIPI & Mid NDVI & Mid ARVI\\
 \hline
 Low TCARI & Low MCARI & Very High ARI1 & V. High ARI2\\
 \hline
\end{tabular}

~
\\(b) Nearest vegetation index bands using \\veg.index bands directly.

\begin{tabular}{|@{}c@{}|@{}c@{}|@{}c@{}|@{}c@{}|}
 \hline
 \textbf{{Veg. Ind. band}} & \textbf{1} & \textbf{2} &\textbf{3}\\
 \hline
 Low EVI & V. High PSRI & Low NDVI & Low ARVI\\
 \hline
 V. High MRENVI & V. High RENDVI & V. High MRESRI & Low VREI2\\
 \hline
 V. High PSRI & Low EVI & Low MCARI2 & Low NDVI\\
 \hline
 Mid SRI & Mid ARVI & Mid NDVI & Mid SIPI\\
 \hline
 Low TCARI & Low MCARI & V. High MRENVI & V. High CRI2\\
 \hline
\end{tabular}
\end{center}
\end{table}

\section{Conclusion}
The paper analyzes how hyperspectral imagery data can be examined using a neural network-driven embedding space. Our experiments demonstrate that such a self-supervised approach has the potential to retrieve unique features different from any non-contextual vector space. In the future, we plan to study ways to model harvest amounts using hyperspectral images of multiple seasons of the same crop field. We aim to design models to predict diseases for early intervention with plant-stress-related data.


\begin{acks}
	This work is partly supported by NSF (Award Number: 1643944) and the USDA Hatch program (Accession Number. 1022633).
\end{acks}

\bibliographystyle{ACM-Reference-Format}
\bibliography{main}

\appendix

\end{document}